%% file: main.tex
\documentclass[conference]{IEEEtran}
\IEEEoverridecommandlockouts
\usepackage{cite}
\usepackage{amsmath,amssymb,amsfonts}
\usepackage{algorithmic}
\usepackage{graphicx}
\usepackage{textcomp}
\usepackage{xcolor}

\usepackage{microtype}
\usepackage{paralist}
\usepackage{subfig}
\usepackage{multirow}
\usepackage{booktabs}
\usepackage{enumitem}

\usepackage{bbm}
\usepackage{amsthm}
\newtheorem{definition}{Definition}

\usepackage[normalem]{ulem}
\useunder{\uline}{\ul}{}

\usepackage{algorithm}
\usepackage{algorithmic}

\DeclareMathOperator*{\argmax}{\arg\!\max}

\graphicspath{{./figs}}

\def\BibTeX{{\rm B\kern-.05em{\sc i\kern-.025em b}\kern-.08em
    T\kern-.1667em\lower.7ex\hbox{E}\kern-.125emX}}
\begin{document}

\title{Distributional Shift Adaptation using Domain-Specific Features}


\author{
    \IEEEauthorblockN{Anique Tahir\IEEEauthorrefmark{1}, 
                    Lu Cheng\IEEEauthorrefmark{2}, 
                    Ruocheng Guo\IEEEauthorrefmark{3} and Huan Liu\IEEEauthorrefmark{1} }
     \IEEEauthorblockA{ 
        \textit{\IEEEauthorrefmark{1}Arizona State University}, 
        Tempe, AZ, USA \\
     } 
     \IEEEauthorblockA{ 
        \textit{\IEEEauthorrefmark{2}University of Illinois Chicago},
        Chicago, IL, USA \\
     } 
     \IEEEauthorblockA{ 
        \textit{\IEEEauthorrefmark{3}Bytedance AI Lab}, 
        London, UK \\
     }
     
 
 }


\maketitle

\begin{abstract}
\input{Sections/abstract}
\end{abstract}

\section{Introduction}
\input{Sections/introduction}

\section{Related Work}
\input{Sections/related_work.tex}

\section{Preliminaries}

\input{Sections/preliminaries.tex}

\section{Method}

\input{Sections/methodology.tex}

\section{Experiments}
\input{Sections/experiments.tex}

\section{Conclusion}
\input{Sections/conclusion.tex}

\section*{Acknowledgements}
We thank Paras Sheth for his valuable help and feedback. This work was supported by the Office of Naval Research under Award No. N00014-21-1-4002. Opinions, interpretations, conclusions, and recommendations are those of the authors.

\bibliographystyle{IEEEtran}
\bibliography{IEEEabrv,main}


\end{document}

%% file: Sections/abstract.tex
Machine learning algorithms typically assume that the training and test samples come from the same distributions, i.e., \textit{in-distribution}.
However, in open-world scenarios, streaming big data can be \textit{Out-Of-Distribution (OOD)}, rendering these algorithms ineffective. Prior solutions to the OOD challenge seek to identify \textit{invariant} features across different training domains. The underlying assumption is that these invariant features should also work reasonably well in the unlabeled target domain. By contrast, this work is interested in the \textit{domain-specific} features that include both invariant features and features unique to the target domain. We propose a simple yet effective approach that relies on correlations in general regardless of whether the features are invariant or not. Our approach uses the most confidently predicted samples identified by an OOD base model (teacher model) to train a new model (student model) that effectively adapts to the target domain. Empirical evaluations on benchmark datasets show that the performance is improved over the SOTA by ${\sim}10$-$20\%$\footnote{https://github.com/aniquetahir/SimprovMinimal}.

%% file: Sections/introduction.tex
\begin{figure}
    \centering
    \includegraphics[width=0.45\textwidth]{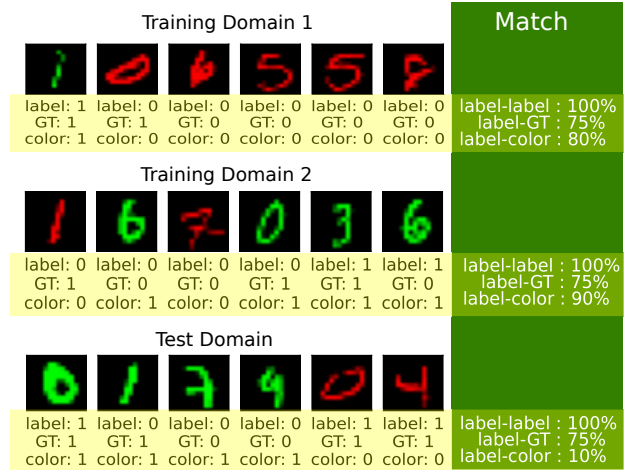}
    \caption{CMNIST dataset sample with color-label and digit-label correlations that vary marginally in the training domains. The test domain has different domain-specific correlations. GT represents the scale of the digit.}
    \label{fig:working_example}
\end{figure}

Standard machine learning models (i.e., models trained by Empirical Risk Minimization (ERM)~\cite{vapnik1991principles}) rely on a key assumption that the training and test data are independent and identically distributed (i.i.d.), or \textit{in-distribution}. However, in practice, streaming big data can be \textit{out-of-distribution (OOD)}, rendering significant performance degradation of ERM-based models. To overcome this critical OOD challenge, a variety of methods have been proposed, such as the Distributionally Robust Optimization (DRO)~\cite{sagawa2019distributionally},
and Invariant Risk Minimization (IRM)~\cite{arjovsky_invariant_2020}. Most of these methods assume that invariant features for prediction across different training domains can also generalize well to the test domain~\cite{li2018domain, edwards2016towards}. However, a comprehensive comparison of different OOD methods by the authors in~\cite{gulrajani_search_2020, schott2021visual} showed that ERM can outperform such methods across different datasets. One potential explanation is that learning invariant features alone may be insufficient.
This work aims to exploit domain-related features to further improve the OOD prediction performance. Take the benchmark dataset CMNIST~\cite{arjovsky_invariant_2020} as an example. In Fig.~\ref{fig:working_example}, we observe that there is a slight difference between the color-label correlation and the digit-label correlation in the training domains. However, the domain-related correlation (color-label) is significantly different between the training and the test domains. This suggests that learning the domain-related features can help predict the label since they capture correlations unique to the test domain.

With the growing popularity of publicly accessible applications and websites, unlabeled data is ubiquitous and contains greater variety, arriving in increasing volumes and with more velocity. Platforms such as Apache Kafka aid in analytics, integrating big data streams. When machine learning models are deployed at scale, the deployment domain might differ from the domains in which the model was trained. Especially, since the model might be trained on a relatively smaller quantity of data compared to the stream of big data it encounters in practice. The incoming unlabeled data, however, might help the model adapt by learning from the distribution of features specific to the deployment domain. This work thereby seeks to leverage \textit{domain-specific} features (including both invariant features and domain-related features) to address the OOD challenge. To achieve this, we assume that the unlabeled data from the target domain is available during deployment for adaptive training.

We identify three primary challenges. First, the latent representations learned during training are often entangled between the invariant features and domain-related features~\cite{zhang2021towards, higgins2016beta, sagawa2019distributionally}. Disentangling these features in the latent space is a challenging task. It is suggested that well-grounded disentanglement approaches must rely on assumptions about the model or data \cite{locatello2019challenging}. Second, how do we identify features related to the target domain with only the labeled training data and unlabeled target data? We need to design a feedback mechanism to enforce the model to learn domain-related features. Finally, with no access to labels for the target domain, it is difficult to determine the optimization direction when adapting the model to the target domain, i.e., determining whether there are positive correlations or negative correlations. This highlights the importance of model selection based on the training data.

To address these challenges, we propose a simple yet effective approach -- \textit{Simprov} -- that learns domain-specific features for OOD prediction using labeled training data and unlabeled target domain data. In particular, we first identify the high confidence predictions in the target domain by using an OOD base model such as IRM and then use these to train a runtime classifier for the target domain. Our major contributions include: (i) a novel framework that uses domain-specific features for OOD prediction, (ii) an effective model selection criterion for fast adapting the model to the target domain, and (iii) empirical analyses on three benchmark datasets from DomainBed~\cite{ganin2016domain} and WILDS~\cite{wilds1, wilds2}.

%% file: Sections/related_work.tex
Standard machine learning uses ERM to optimize the objective function. A key assumption is that random variables in the data are \textit{i.i.d}. Thus, in scenarios involving distribution shift, ERM performance degrades significantly~\cite{lazer2014parable, rosenfeld2022online}. OOD methods aim to address the issue by using data from related domains that differ in distributions.

One seminal work in OOD is Invariant Risk Minimization (IRM)~\cite{arjovsky_invariant_2020} which aims to identify the invariant features. The hypothesis is that if the model can identify the causes (i.e., the invariant features) of an outcome, then it should perform reasonably well in a new unlabeled domain as it does not rely on spurious correlations. Distributionally Robust Optimization (DRO) family of approaches focuses on the worst-case scenario~\cite{rahimian2019distributionally, hu2018does}: optimizing for the source domain with the greatest loss. Another line of research leverages pseudo-labeling and data augmentation~\cite{sohn2020fixmatch, cubuk2020randaugment} approaches. Here, a trained model is used to generate noisy labels for samples in the unlabeled domain, combine them with the annotated training data and use the resulting semi-pseudo-labeled batch to further improve the trained model~\cite{berthelot2019mixmatch}. Noisy student~\cite{xie2020self} incorporates model distillation, where it trains the teacher to generate pseudo labels which are used for training a student model.

More recent research~\cite{long2017deep,ganin2016domain} considered Domain Adaptation (DA), where both the labeled training domains and unlabeled test domain(s) are available during training. Our problem setting is slightly different: we optimize prediction performance while DA optimizes on the learned representation.
Adaptive Risk Minimization (ARM)~\cite{zhang_adaptive_2020}
studied the same problem setting as ours by adapting the training model to the target domain using meta-learning to update the model's parameters. 
We complement prior works by considering the importance of learning domain-specific features for the target domain and removing potential spurious features identified in the training domains, that is, features useful for prediction during training but not for the target domain. 


%% file: Sections/preliminaries.tex
\textbf{Invariant Risk Minimization.} IRM~\cite{arjovsky_invariant_2020} aims to identify the invariant features (often referred to as \textit{causes}) by training over multiple different domains. 
Thus the loss function of IRM is designed to minimize the per domain risk, $R^e = \mathbb{E}_{p^{tr}(x,y|e)}[l]$, where $x$ represents the features, $y$ the labels, $e$ the domain, $l$ the loss function (e.g., mean squared error), and $p^{tr}$ is the distribution over the training domains. Formally, let $\Phi$ be the invariant prediction function. The objective function of IRM, $L$, can be then defined as:
\begin{equation}
    L(\Phi) = \sum_{e \in \mathcal{E}^{tr}} R^e(\Phi) + \lambda|| \nabla_{\hat{w}|\hat{w}=1.0} R^e (\hat{w} \circ \Phi) ||,
\end{equation}
where $\hat{w}$ is a classification model that predicts from the invariant features, $\lambda$ is the regularization parameter, and $\mathcal{E}^{tr}$ is the set of training domains. The second term adds a constraint on the learning for a particular environment by increasing the loss when the propagation gradients are high resulting in reduced learning towards a specific domain leading to more generalizability.
\\\textbf{Distillation.}
The distillation consists of a teacher model and a student model. The teacher model is trained on the original data and the student model then learns from the teacher~\cite{sanh2019distilbert, hinton2015distilling}. 
In this work, we use offline and response-based distillation~\cite{gou2021knowledge} where the predictions (hard-labels) or logits (soft-labels) of the teacher model are used to train the student model.

Formally, let $T$ denote the teacher model, $S$ the student model, the recursive loss function $L$ for offline distillation is:
\begin{equation}
    L = \sum_{i=1}^{n}\alpha L(y_i, S(x_i)) + (1-\alpha)L(y_i, T(x_i)),
\end{equation}
where $\alpha$ denotes the ratio between the two losses~(teachers and students) and $n$ is the number of samples. Here, the teacher and student models are trained independently from each other.



%% file: Sections/methodology.tex
In this section, we describe our proposed approach~(\textit{Simprov}) for tackling the OOD challenge. 
Simprov aims to effectively adapt an ERM-based model to the distribution of target domain by learning domain-specific features. We first formally define the problem setting as follows:
\begin{definition}
\label{def:arm_setting}
Let $\mathcal{E}_{all}$ be the set of all possible domains, $\mathcal{E}_{tr}$ the set of training domains, and $\mathcal{E}_{te}$ the set containing the target domain. Given training samples $x_{tr} \in \mathcal{X}$ of an input random variable X, $y_{tr} \in \mathcal{Y}$ of a target random variable Y, $z \in \mathcal{Z}$ of an input domain random variable Z , and $x_{te} \in \mathcal{X}$ of X, the goal is to learn a function $f: \mathcal{X} \rightarrow \mathcal{Y}$ representing $P(Y|X, e_{te})$ given, $P(X, Y|e_{te}) \neq P(X,Y|e_{tr})$, where  $e_{te} \in \mathcal{E}_{te}$ and $e_{tr} \in \mathcal{E}_{tr}$. 
\end{definition}
\begin{figure}
    \centering
    \includegraphics[width=0.4\textwidth]{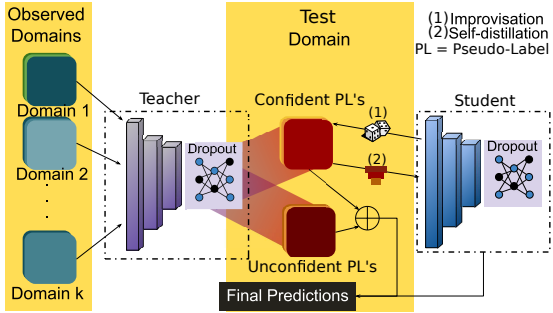}
    \caption{An overview of Simprov. It leverages invariant features (Teacher), distillation, and a model selection criterion (Improvisation) to enhance performance on the target domain. The teacher model (such as IRM) learns invariant features and identifies target samples predicted with high confidence. Pseudo-labels (PL) with dropout are used to estimate the confidence. The student model (i.e., an ERM-based model) is trained over these selected samples to make predictions in the target domain. Since the pseudo-labels are generated without prior knowledge of the target domain, training over them requires a positive feedback loop between the teacher and student formed by the combination of Improvisation (1) and Self-Distillation (2).}
    \label{fig:architecture}
\end{figure}
\raggedbottom

\sloppy An overview of our approach is highlighted in Fig.~\ref{fig:architecture}. Simprov primarily consists of three parts: Pseudo-Labeling, Self-Distillation, and Model Selection (Improvisation). Logically, the structure of the prediction model constitutes of a representation learning module $\Phi: \mathcal{X} \rightarrow \mathcal{H}$ and a classifier $\hat{w}: \mathcal{H} \rightarrow \mathcal{Y}$, where $\mathcal{H}$ is the representation space. We aim to learn the function $f: \mathcal{X} \rightarrow \mathcal{Y} \sim P(Y|X, e_{te})$. Note the difference between our problem setting and Domain Adaptation is that the objective of the latter is to learn the invariant representation function $\Phi$ s.t. $P(\Phi(X)|X, e_{te}) = P(\Phi(X)|X, e_{tr})$.

\subsection{Pseudo-Labeling}
Simprov's learning is initiated by pseudo-labeling the target data. It then relies on a positive feedback loop for learning about the target distribution. Simprov first identifies the subset of the high confidence target predictions using a base model for OOD generalization. The intuition is that the predictions with the highest confidence are the most accurate since the confidence represents the reliance on invariant features for the predictions.
Simprov then uses these predictions with high confidence as pseudo-labels to start a feedback loop.

Particularly, we use a trained OOD base model such as IRM to pseudo-label target data with prediction confidence values generated using Monte Carlo (MC) dropout for uncertainty estimation~\cite{gal2016dropout}. Specifically, we perform label inference after changing the dropout mask for the same batch of target data. The variance between the inferences then determines the confidence in the predictions. Formally, let $\mathcal{C}=\{1, 2, ..., k\}$ be the set of $k$ classes, $d$ the dropout probability, $m$ the number of confirmations for the pseudo-labeling process, and $f$ the labeling function parameterized by $\theta$. The pseudo-label $\Tilde{l}_i$ of the $i$-th inference for target sample $j$ can be obtained by:
\begin{equation}
    \Tilde{l}_{j,i} = f_\theta(x_j, d_{j,i})\quad \forall i \in \{1, 2, ..., m\}, x_j \in \mathcal{E}_{te}.
\end{equation}
\sloppy Let $\ell_j$ be the set of all inferred pseudo labels of $j$ i.e., $\ell_j = \{\tilde{l}_{j, 1}, \tilde{l}_{j, 2}, ..., \tilde{l}_{j, m}\}$. A simple majority voting strategy is used to infer the final pseudo label $\Tilde{y}_j$:
\begin{equation}
    \label{eq:mode}
    \Tilde{y}_j = \argmax_c \sum_{a} \mathbbm{1}(a, c) \quad \forall a \in \ell_j, c \in \mathcal{C}.
\end{equation}
where $\mathbbm{1}$ is the indicator function. Finally, the confidence score $\kappa_j$ of $\Tilde{y}_j$ is defined as $\kappa_j = - \text{Var} (\ell_j)$, i.e., the variance of $\ell_j$.



\subsection{Self-Distillation}

With the high confidence target samples predicted by an OOD base model, Simprov trains an ERM-based student model with dropouts over the target distribution. To further improve the quality of the pseudo-labels,  it creates a positive feedback loop where it re-trains the student model using the previous student model as the teacher.
At the end of the feedback loop, Simprov learns domain-specific features in the target domain. 


At t=0, $f_{\theta_0}$ is the function learned by the base model (e.g., IRM on training domains). It encourages Simprov to predict using invariant features. $f_{\theta_0}$ is then used to infer pseudo labels for the target data. Next, we train the student model $f_{\theta_1}$ on the target data using the pseudo labels. We update the pseudo labels for the target data using re-trained $f_{\theta_1}$. The iterative process improves the student model towards positive feedback as judged by the model selection criterion detailed below. 
\subsection{Random Chance-based Model Selection}
Although the self-distillation process can help improve the quality of the pseudo-labels, it might turn into a negative feedback loop as the correct direction of the feedback loop is unknown. Incorrect pseudo labels will only reinforce the teacher's inconsistencies.

\sloppy To address this challenge, we propose to use the student model's pseudo-labels for training domains to maneuver the direction of the feedback loop in the self-distillation process. 
The base model learns invariant features in the training domains. However, due to issues such as sufficiency~\cite{lee2017sufficiency}, it may learn some spurious features. Between training and target domains, these spurious features ($X_{spur}$) may be 
(i) positively correlated i.e., $P(Y|X_{spur}, e_{te}) \propto P(Y|X_{spur}, e_{tr})$, (ii) negatively correlated i.e., $P(Y|X_{spur}, e_{te}) \propto \frac{1}{P(Y|X_{spur}, e_{tr})}$, or (iii) independent. By definition, $P(Y|X_{inv},e_{tr})=P(Y|X_{inv},e_{ts})$, where $X_{inv}$ represents the invariant latent features. If the training and target distributions have the same correlation~(cases (i) and (iii)), then a model trained on the target distribution works similarly on the training distribution. Otherwise~(case (ii)),
the model would give an accuracy that is lower than random chance on $\mathcal{E}_{tr}$.
We propose a new metric $d^t_{rand}$ to help identify the direction of the self-distillation feedback loop: the difference between the training prediction accuracy and the random chance of a model trained on target pseudo-labels. Formally, the model selection metric is defined as:
\begin{equation}
    d^t_{rand} = \Big\lvert f_{\theta_t}(x) - \frac{1}{k}\Big\rvert \ ,\quad x \in \mathcal{E}_{tr}.
\end{equation}

If $d^t_{rand}$ is greater than $d^{t-1}_{rand}$, it indicates that the model has learned informative features. Thus, during self-distillation, Simprov only replaces the teacher model at $t-1$ with the student model when this metric increases. This ensures that there is information gain from the target distribution to de-noise the pseudo-labels, i.e., the model is learning the domain-specific features, including both the domain-relevant and invariant features in the target domain.

%% file: Sections/experiments.tex
We aim to answer the following research questions in the experiments: \textbf{RQ. 1} Can Simprov outperform SOTA for OOD over different datasets? \textbf{RQ. 2} How effective is the proposed model selection criterion? \textbf{RQ. 3} How sensitive is Simprov to different values of hyperparameters?

\begin{table}[]
\centering
\resizebox{\columnwidth}{!}{
\begin{tabular}{llll}
\cline{2-4}
                            & \textbf{CMNIST}      & \textbf{Camelyon17}  & \textbf{Waterbirds} \\ \cline{2-4} 
IRM                         & {\ul 67.1 (2.5)}           & 64.2 (8.1)           & 75.3 (0.6)          \\
Group DRO                   & 38.7 (1.8)           & 68.4 (7.3)           & 91.4 (0.3)          \\ \hline
DANN                        & 51.5 (0.3)           & 68.4 (9.2)           & 77.8 (0.0)          \\ 
ARM                         & 56.2 (0.2)     & 87.2 (0.9)     & {\ul 94.1 (0.0)}    \\
Pseudolabel                 & 42.9 (1.1)           & 67.7 (8.2)           & 74.2 (8.0)          \\
NoisyStudent                & 27.1 (3.8)           & 86.7 (1.7)           & 22.2 (0.0)               \\ \hline
\textbf{Simprov-IRM (Ours)} & \textbf{89.8 (0.1)} & \textbf{92.8 (6.2)} & 81.6 (8.1)         \\
\textbf{Simprov-DRO (Ours)} & 12.3 (0.0)           & {\ul 87.7 (3.5)}                & \textbf{95.0 (3.0)}
\end{tabular}
}
\caption{Average accuracy and standard deviations over five trials of different methods under three benchmark datasets.}
\label{tbl:accuracies}
\end{table}

\subsection{Experimental Setup}
\label{sec:setup}
Our implementation extends the boilerplate provided by the Stanford's WILDS benchmark repository~\cite{wilds2}.

\textit{Datasets.} We use three benchmark datasets with different classification tasks. (i) CMNIST~\cite{arjovsky_invariant_2020} contains images of digits that have either of the two colors: green and red. The label is `1’ if the digit is less than five, otherwise it is `0’. (ii) Camelyon17-Wilds~\cite{bandi2018detection} is related to tumor detection. (iii) Waterbirds~\cite{sagawa2019distributionally} aim to classify images of landbirds and waterbirds with land or water backgrounds. For the model architecture, we followed the default setting of WILDS~\cite{wilds1}.

\textit{Baselines}. We compare Simprov with two popular OOD models (i.e., IRM~\cite{arjovsky_invariant_2020} and Group DRO~\cite{rahimian2019distributionally}) and four SOTA domain-adaptation models (i.e., DANN~\cite{ganin2016domain}, ARM~\cite{zhang_adaptive_2020}, Pseudolabel~\cite{lee2013pseudo}, and NoisyStudent~\cite{xie2020self}). IRM and Group DRO aim to learn invariant features across domains. DANN, ARM, Pseudolabel, and NoisyStudent employ techniques to ensure the distributions of learned representations are aligned across domain. Using IRM and DRO base models leads to two versions of Simprov: \textit{Simprov-IRM} and \textit{Simprov-DRO}.

\subsection{Results}
\label{sec:results}
We report the mean and standard deviations of the accuracy on the target domain over five trials of the selected models. We present the results in Table~\ref{tbl:accuracies}. We used the same train/test splits (i.e., the hardest case) for CMNIST as in~\cite{arjovsky_invariant_2020}, different from most of other implementations that report results over a combination of splits. The best results are in bold font and the second best ones are underlined.
We make the following observations answering \textbf{RQ1}:
\begin{itemize}[leftmargin=*]
    \item Simprov mostly outperforms the corresponding base models across different tasks (e.g., Simprov-IRM outperforms IRM for CMNIST), indicating that learning domain-specific features is critical for achieving high accuracy in OOD tasks. Simprov improves accuracy by ${\sim}20\%$ on the hardest dataset~(CMNIST) as it optimizes the feature representation using target domain data.
    \item Simprov reinforces the feature correlations learned in the base models. This is supported by the observation that when the base models perform relatively well (e.g., Camelyon17 and Waterbirds), it can improve the prediction performance by a large margin; however, its performance degrades significantly if the base models fail (e.g., Group DRO for CMNIST). This further implies that learning invariant features is necessary for the OOD challenge.
    \item Compared to the SOTA models for domain adaption, our models consistently achieve the best performance. For example, Simprov has an ${\sim}10\%$ improvement on average over three datasets on compared to ARM. There are two reasons for this improvement. First, by using only the pseudo-labels predicted by the OOD base models rather than their latent features, 
    By using pseudo labels instead of the features for the training data during distillation, Simprov does not rely on the strong feedback regarding the training domains while retaining feedback for the target domain via backpropagation; (ii) Our model selection strategy leads the training process in a direction of information gain, i.e., when the random-chance difference is large, Simprov has high information about the target domain, leading to comparatively better performance.
\end{itemize}

\begin{figure}[]%
    \centering
    \subfloat[\centering $D$]{{\includegraphics[width=0.21\textwidth]{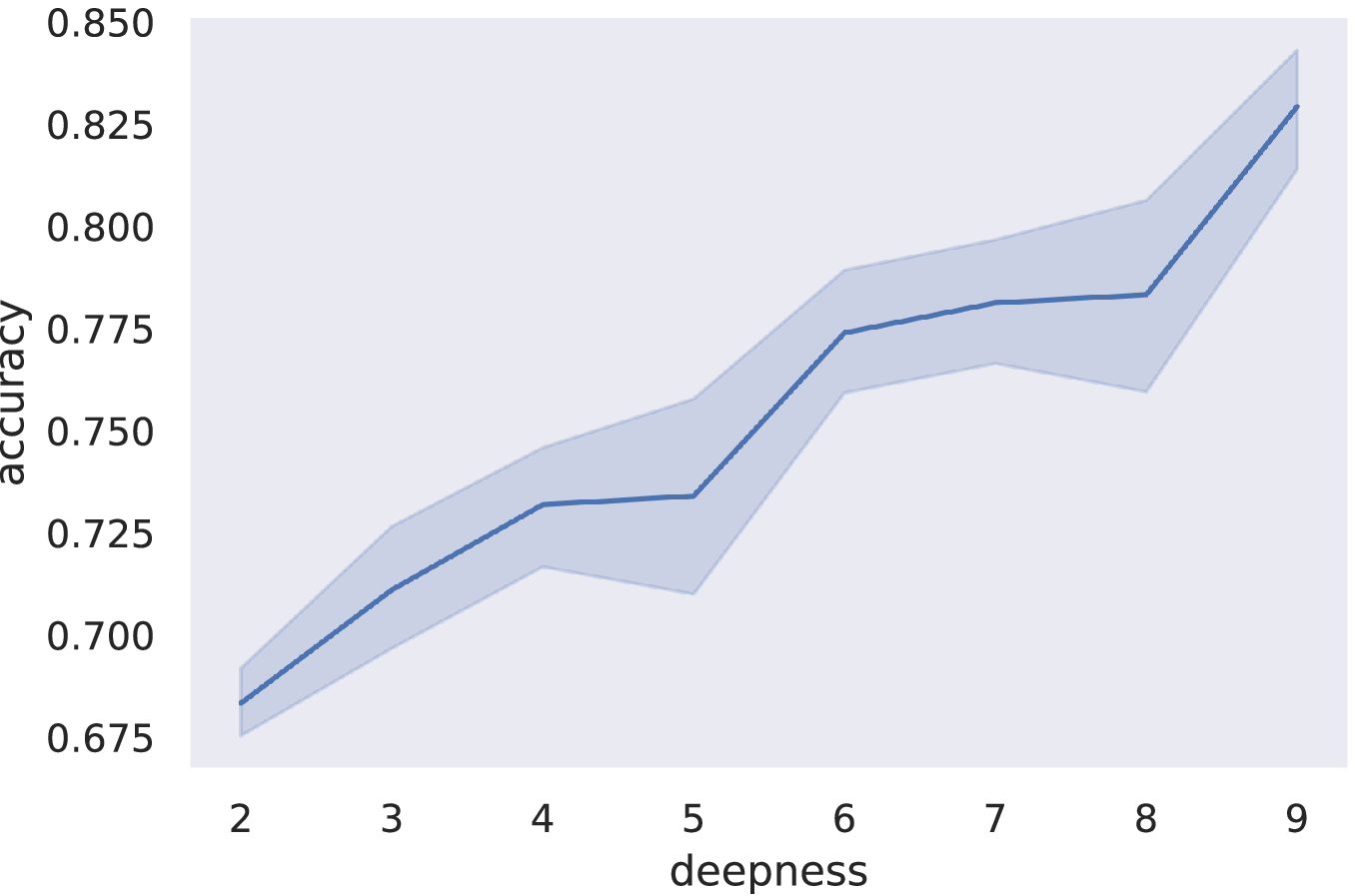} }}%
    \qquad
    \subfloat[\centering $d^t_{rand}$]{{\includegraphics[width=0.21\textwidth]{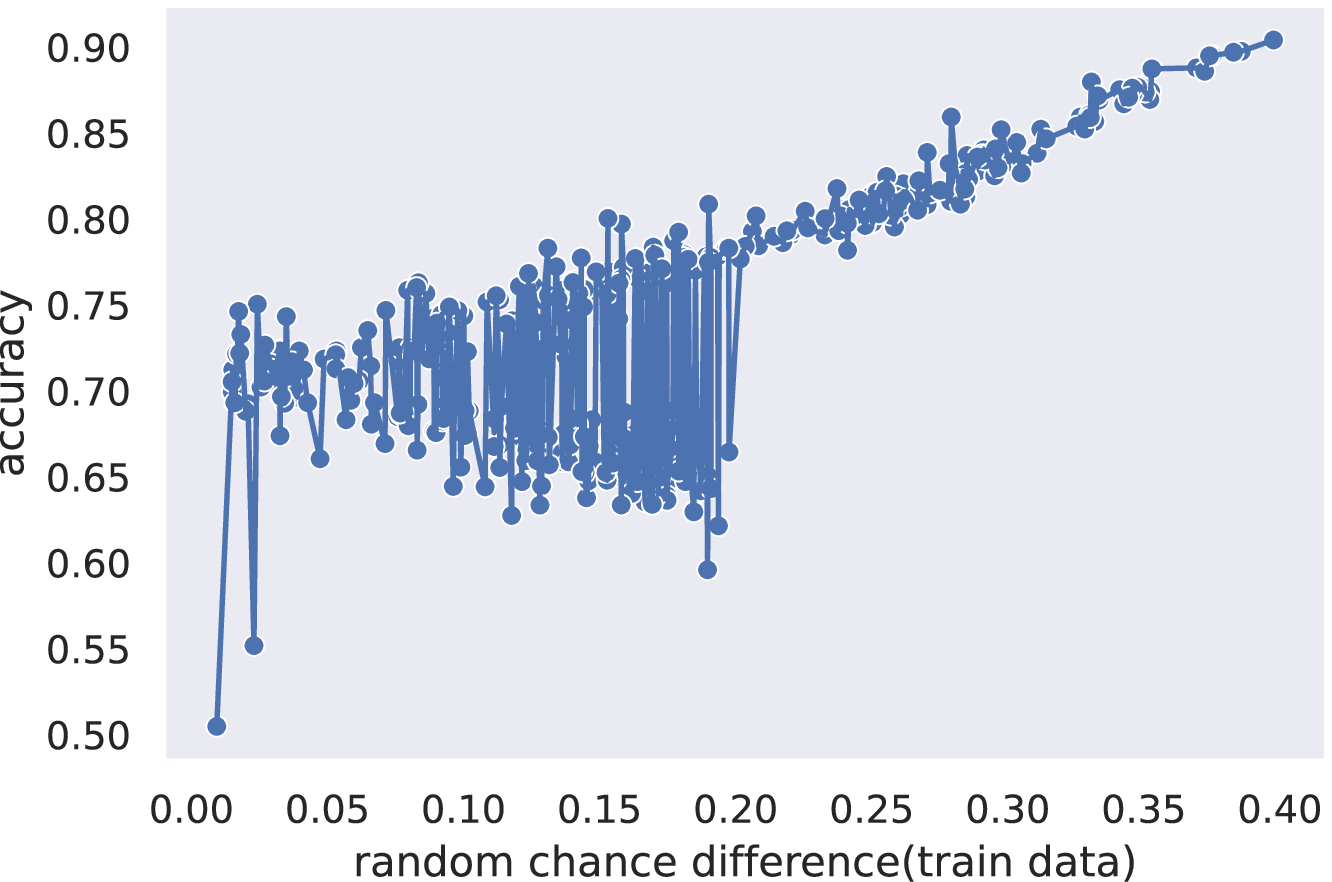} }}%
    \caption{(a) effects of deepness on the accuracy, and (b) accuracy changes relative to our model selection metric.}%
    \label{fig:model_selection}%
\end{figure}

We perform further analysis to answer \textbf{RQ2} and \textbf{RQ3}. Fig.~\ref{fig:model_selection}(a) shows that increasing the deepness (i.e. the number of distillation iterations) of the self-distillation process generally helps Simprov learn better domain-specific features in the target domain. We believe this is in part due to the feedback loop created during training on the target data. Fig.~\ref{fig:model_selection}(b) shows that the proposed model selection strategy is effective. When the random chance difference is low, there is high variation in the accuracy of the models on the target domain. This is because the closer the model's performance to random chance accuracy on the training data, the less Simprov knows about the domain-specific features. By contrast, at a higher random chance difference, Simprov presents more minor variations and higher accuracy on the target domain.

%% file: Sections/conclusion.tex
Our approach~(Simprov) leveraged both labeled training data and target data to learn domain-specific features guided by an effective model selection criterion. We showed that our method can outperform SOTA over three benchmark datasets. We draw two main conclusions. First, our approach relies on invariants from OOD models in prior works. Second, our approach does not find \textit{purely} invariant features in the data in lieu of the domain-specific features. We leave these to future work.


%% file: main.bbl
\begin{thebibliography}{10}
\providecommand{\url}[1]{#1}
\csname url@samestyle\endcsname
\providecommand{\newblock}{\relax}
\providecommand{\bibinfo}[2]{#2}
\providecommand{\BIBentrySTDinterwordspacing}{\spaceskip=0pt\relax}
\providecommand{\BIBentryALTinterwordstretchfactor}{4}
\providecommand{\BIBentryALTinterwordspacing}{\spaceskip=\fontdimen2\font plus
\BIBentryALTinterwordstretchfactor\fontdimen3\font minus
  \fontdimen4\font\relax}
\providecommand{\BIBforeignlanguage}[2]{{%
\expandafter\ifx\csname l@#1\endcsname\relax
\typeout{** WARNING: IEEEtran.bst: No hyphenation pattern has been}%
\typeout{** loaded for the language `#1'. Using the pattern for}%
\typeout{** the default language instead.}%
\else
\language=\csname l@#1\endcsname
\fi
#2}}
\providecommand{\BIBdecl}{\relax}
\BIBdecl

\bibitem{vapnik1991principles}
V.~Vapnik, ``Principles of risk minimization for learning theory,''
  \emph{Advances in neural information processing systems}, vol.~4, 1991.

\bibitem{sagawa2019distributionally}
S.~Sagawa, P.~W. Koh, T.~B. Hashimoto, and P.~Liang, ``Distributionally robust
  neural networks for group shifts: On the importance of regularization for
  worst-case generalization,'' \emph{arXiv preprint arXiv:1911.08731}, 2019.

\bibitem{arjovsky_invariant_2020}
\BIBentryALTinterwordspacing
M.~Arjovsky, L.~Bottou, I.~Gulrajani, and D.~Lopez-Paz, ``Invariant {Risk}
  {Minimization},'' \emph{arXiv:1907.02893 [cs, stat]}, Mar. 2020, arXiv:
  1907.02893. [Online]. Available: \url{http://arxiv.org/abs/1907.02893}
\BIBentrySTDinterwordspacing

\bibitem{li2018domain}
H.~Li, S.~J. Pan, S.~Wang, and A.~C. Kot, ``Domain generalization with
  adversarial feature learning,'' in \emph{2018 IEEE/CVF Conference on Computer
  Vision and Pattern Recognition}.\hskip 1em plus 0.5em minus 0.4em\relax IEEE,
  2018, pp. 5400--5409.

\bibitem{edwards2016towards}
H.~Edwards and A.~Storkey, ``Towards a neural statistician,'' \emph{arXiv
  preprint arXiv:1606.02185}, 2016.

\bibitem{gulrajani_search_2020}
\BIBentryALTinterwordspacing
I.~Gulrajani and D.~Lopez-Paz, ``In {Search} of {Lost} {Domain}
  {Generalization},'' \emph{arXiv:2007.01434 [cs, stat]}, Jul. 2020. [Online].
  Available: \url{http://arxiv.org/abs/2007.01434}
\BIBentrySTDinterwordspacing

\bibitem{schott2021visual}
L.~Schott, J.~von K{\"u}gelgen, F.~Tr{\"a}uble, P.~Gehler, C.~Russell,
  M.~Bethge, B.~Sch{\"o}lkopf, F.~Locatello, and W.~Brendel, ``Visual
  representation learning does not generalize strongly within the same
  domain,'' \emph{arXiv preprint arXiv:2107.08221}, 2021.

\bibitem{zhang2021towards}
H.~Zhang, Y.-F. Zhang, W.~Liu, A.~Weller, B.~Sch{\"o}lkopf, and E.~P. Xing,
  ``Towards principled disentanglement for domain generalization,'' \emph{arXiv
  preprint arXiv:2111.13839}, 2021.

\bibitem{higgins2016beta}
I.~Higgins, L.~Matthey, A.~Pal, C.~Burgess, X.~Glorot, M.~Botvinick,
  S.~Mohamed, and A.~Lerchner, ``beta-vae: Learning basic visual concepts with
  a constrained variational framework,'' 2016.

\bibitem{locatello2019challenging}
F.~Locatello, S.~Bauer, M.~Lucic, G.~Raetsch, S.~Gelly, B.~Sch{\"o}lkopf, and
  O.~Bachem, ``Challenging common assumptions in the unsupervised learning of
  disentangled representations,'' in \emph{international conference on machine
  learning}.\hskip 1em plus 0.5em minus 0.4em\relax PMLR, 2019, pp. 4114--4124.

\bibitem{ganin2016domain}
Y.~Ganin, E.~Ustinova, H.~Ajakan, P.~Germain, H.~Larochelle, F.~Laviolette,
  M.~Marchand, and V.~Lempitsky, ``Domain-adversarial training of neural
  networks,'' \emph{The journal of machine learning research}, vol.~17, no.~1,
  pp. 2096--2030, 2016.

\bibitem{wilds1}
\BIBentryALTinterwordspacing
P.~W. Koh, S.~Sagawa, H.~Marklund, S.~M. Xie, M.~Zhang, A.~Balsubramani, W.~Hu,
  M.~Yasunaga, R.~L. Phillips, S.~Beery, J.~Leskovec, A.~Kundaje, E.~Pierson,
  S.~Levine, C.~Finn, and P.~Liang, ``{WILDS:} {A} benchmark of in-the-wild
  distribution shifts,'' \emph{CoRR}, vol. abs/2012.07421, 2020. [Online].
  Available: \url{https://arxiv.org/abs/2012.07421}
\BIBentrySTDinterwordspacing

\bibitem{wilds2}
\BIBentryALTinterwordspacing
S.~Sagawa, P.~W. Koh, T.~Lee, I.~Gao, S.~M. Xie, K.~Shen, A.~Kumar, W.~Hu,
  M.~Yasunaga, H.~Marklund, S.~Beery, E.~David, I.~Stavness, W.~Guo,
  J.~Leskovec, K.~Saenko, T.~Hashimoto, S.~Levine, C.~Finn, and P.~Liang,
  ``Extending the {WILDS} benchmark for unsupervised adaptation,'' \emph{CoRR},
  vol. abs/2112.05090, 2021. [Online]. Available:
  \url{https://arxiv.org/abs/2112.05090}
\BIBentrySTDinterwordspacing

\bibitem{lazer2014parable}
D.~Lazer, R.~Kennedy, G.~King, and A.~Vespignani, ``The parable of google flu:
  traps in big data analysis,'' \emph{science}, vol. 343, no. 6176, pp.
  1203--1205, 2014.

\bibitem{rosenfeld2022online}
E.~Rosenfeld, P.~Ravikumar, and A.~Risteski, ``An online learning approach to
  interpolation and extrapolation in domain generalization,'' in
  \emph{International Conference on Artificial Intelligence and
  Statistics}.\hskip 1em plus 0.5em minus 0.4em\relax PMLR, 2022, pp.
  2641--2657.

\bibitem{rahimian2019distributionally}
H.~Rahimian and S.~Mehrotra, ``Distributionally robust optimization: A
  review,'' \emph{arXiv preprint arXiv:1908.05659}, 2019.

\bibitem{hu2018does}
W.~Hu, G.~Niu, I.~Sato, and M.~Sugiyama, ``Does distributionally robust
  supervised learning give robust classifiers?'' in \emph{International
  Conference on Machine Learning}.\hskip 1em plus 0.5em minus 0.4em\relax PMLR,
  2018, pp. 2029--2037.

\bibitem{sohn2020fixmatch}
K.~Sohn, D.~Berthelot, N.~Carlini, Z.~Zhang, H.~Zhang, C.~A. Raffel, E.~D.
  Cubuk, A.~Kurakin, and C.-L. Li, ``Fixmatch: Simplifying semi-supervised
  learning with consistency and confidence,'' \emph{Advances in Neural
  Information Processing Systems}, vol.~33, pp. 596--608, 2020.

\bibitem{cubuk2020randaugment}
E.~D. Cubuk, B.~Zoph, J.~Shlens, and Q.~V. Le, ``Randaugment: Practical
  automated data augmentation with a reduced search space,'' in
  \emph{Proceedings of the IEEE/CVF Conference on Computer Vision and Pattern
  Recognition Workshops}, 2020, pp. 702--703.

\bibitem{berthelot2019mixmatch}
D.~Berthelot, N.~Carlini, I.~Goodfellow, N.~Papernot, A.~Oliver, and C.~A.
  Raffel, ``Mixmatch: A holistic approach to semi-supervised learning,''
  \emph{Advances in Neural Information Processing Systems}, vol.~32, 2019.

\bibitem{xie2020self}
Q.~Xie, M.-T. Luong, E.~Hovy, and Q.~V. Le, ``Self-training with noisy student
  improves imagenet classification,'' in \emph{Proceedings of the IEEE/CVF
  conference on computer vision and pattern recognition}, 2020, pp.
  10\,687--10\,698.

\bibitem{long2017deep}
M.~Long, H.~Zhu, J.~Wang, and M.~I. Jordan, ``Deep transfer learning with joint
  adaptation networks,'' in \emph{International conference on machine
  learning}.\hskip 1em plus 0.5em minus 0.4em\relax PMLR, 2017, pp. 2208--2217.

\bibitem{zhang_adaptive_2020}
M.~M. Zhang, H.~Marklund, N.~Dhawan, A.~Gupta, S.~Levine, and C.~Finn,
  ``Adaptive risk minimization: {A} meta-learning approach for tackling group
  shift,'' 2020.

\bibitem{sanh2019distilbert}
V.~Sanh, L.~Debut, J.~Chaumond, and T.~Wolf, ``Distilbert, a distilled version
  of bert: smaller, faster, cheaper and lighter,'' \emph{arXiv preprint
  arXiv:1910.01108}, 2019.

\bibitem{hinton2015distilling}
G.~Hinton, O.~Vinyals, J.~Dean \emph{et~al.}, ``Distilling the knowledge in a
  neural network,'' \emph{arXiv preprint arXiv:1503.02531}, vol.~2, no.~7,
  2015.

\bibitem{gou2021knowledge}
J.~Gou, B.~Yu, S.~J. Maybank, and D.~Tao, ``Knowledge distillation: A survey,''
  \emph{International Journal of Computer Vision}, vol. 129, no.~6, pp.
  1789--1819, 2021.

\bibitem{gal2016dropout}
Y.~Gal and Z.~Ghahramani, ``Dropout as a bayesian approximation: Representing
  model uncertainty in deep learning,'' in \emph{international conference on
  machine learning}.\hskip 1em plus 0.5em minus 0.4em\relax PMLR, 2016, pp.
  1050--1059.

\bibitem{lee2017sufficiency}
J.~Y. Lee, J.~J. Brown, and L.~M. Ryan, ``Sufficiency revisited: Rethinking
  statistical algorithms in the big data era,'' \emph{The American
  Statistician}, vol.~71, no.~3, pp. 202--208, 2017.

\bibitem{bandi2018detection}
P.~Bandi, O.~Geessink, Q.~Manson, M.~Van~Dijk, M.~Balkenhol, M.~Hermsen, B.~E.
  Bejnordi, B.~Lee, K.~Paeng, A.~Zhong \emph{et~al.}, ``From detection of
  individual metastases to classification of lymph node status at the patient
  level: the camelyon17 challenge,'' \emph{IEEE Transactions on Medical
  Imaging}, 2018.

\bibitem{lee2013pseudo}
D.-H. Lee \emph{et~al.}, ``Pseudo-label: The simple and efficient
  semi-supervised learning method for deep neural networks,'' in \emph{Workshop
  on challenges in representation learning, ICML}, vol.~3, no.~2, 2013, p. 896.

\end{thebibliography}
